\pdfoutput=1
\documentclass[letterpaper, 10 pt, conference]{ieeeconf}  

\IEEEoverridecommandlockouts                              

\usepackage{graphics} 
\usepackage{epsfig} 
\usepackage{times} 
\usepackage{amsmath} 
\usepackage{amssymb}  
\usepackage{tabularx}
\usepackage[ruled,vlined]{algorithm2e}
\usepackage{booktabs,multirow,makecell,subcaption}
\usepackage{tabularx}
\usepackage{placeins}
\usepackage{float}
\usepackage[dvipsnames]{xcolor}
\usepackage{threeparttable}
\usepackage[table]{xcolor} 
\usepackage{url}
\usepackage{times}
\title{DexHiL: A Human-in-the-Loop Framework for Vision-Language-Action Model Post-Training in Dexterous Manipulation}

\author{
    Yifan Han$^{*,1}$ \quad Zhongxi Chen$^{*,2}$ \quad Yuxuan Zhao$^2$ \quad Congsheng Xu$^2$ \\
    Yanming Shao$^3$ \quad Yichuan Peng$^2$ \quad Yao Mu$^{2,\dagger}$ \quad Wenzhao Lian$^{2,\dagger}$ \\
    \\
    \small $^1$CASIA \quad $^2$SJTU \quad $^3$Shanghai AI Laboratory \\
    \small $^*$Equal contribution \quad $^\dagger$Corresponding author
}

\begin{document}

\maketitle
\thispagestyle{empty}
\pagestyle{empty}

\begin{abstract}
While Vision-Language-Action (VLA) model has demonstrated promising generalization capabilities in robotic manipulation, deploying them on specific and complex downstream tasks still demands effective post-training. In parallel, Human-in-the-Loop (HiL) learning has proven to be a powerful mechanism for refining robot policies. However, extending this paradigm to dexterous manipulation remains challenging: multi-finger control is high-dimensional, contact-intensive, and exhibits execution distributions that differ markedly from arm motion, leaving existing dexterous VLA systems limited in reliability and adaptability. We present DexHiL, the first integrated arm–hand human-in-the-loop framework for dexterous manipulation VLA , enabling coordinated interventions over the arm and the dexterous hand within a single system. DexHiL introduces an intervention-aware data sampling strategy that prioritizes corrective segments for post-training, together with a lightweight teleoperation interface that supports instant human corrections during execution. Real-robot experiments demonstrate that DexHiL serves as an effective post-training framework, yielding a substantial performance leap that outperforming standard offline-only finetuning baselines by an average of 25\% in success rates across distinct tasks.
Project page: \url{https://chenzhongxi-sjtu.github.io/dexhil/}
\end{abstract}
\section{Introduction}
Vision-Language-Action (VLA) models represent a paradigm shift in robot learning, enabling cross-scenario semantic understanding and spatial perception by directly mapping multimodal inputs to control outputs~\cite{brohan2022rt1,zitkovich2023rt2,kim2024openvla,black2024pi0,liu2024foam}. While VLA models have demonstrated significant potential in general manipulation, their application to high-degree-of-freedom (DOF) dexterous hands remains challenging, particularly in the post-training and downstream adaptation phases. Existing VLA post-training strategies, which typically rely on Supervised Fine-Tuning (SFT) over offline datasets~\cite{kim2024openvla,octo2024,li2024roboflamingo}, struggle to bridge the gap between high-dimensional end-effector control and the intricate, contact-rich requirements of multi-fingered manipulation.

The primary bottleneck in adapting VLA models to dexterous tasks partially stems from hardware-level kinematic misalignment. Traditional teleoperation interfaces, such as exoskeletons and leader-follower arms~\cite{zhao2023aloha,wu2024gello,yang2024ace}, often fail to precisely map human hand motions to the complex joint configurations of robotic hands. This misalignment, coupled with the sharp discontinuities inherent in dexterous contact states, precludes the collection of high-quality, fine-grained demonstration data. Consequently, existing VLA paradigms lack the high-fidelity guidance necessary to optimize the nuanced motion manifold required for robust dexterous performance.
\begin{figure}
    \centering
    \includegraphics[width=1.0\linewidth]{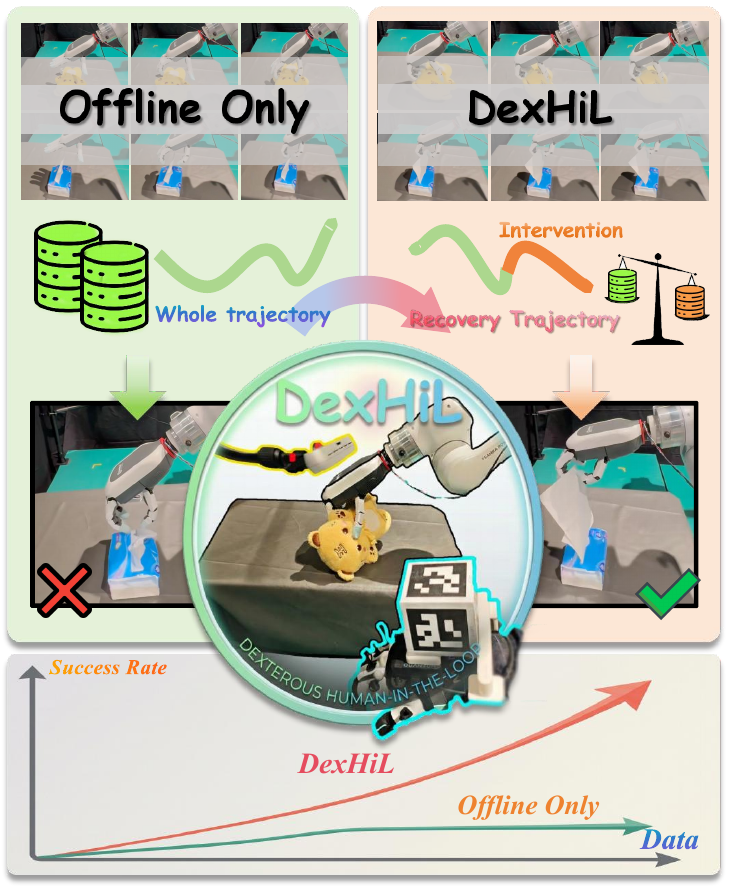}
    \caption{While scaling offline data for VLA models yields slow accuracy gains and performance plateaus, DexHiL integrates offline training with online Human-in-the-Loop interventions. By strategically reweighting offline and corrective online data, our approach achieves high data efficiency and rapid accuracy growth.}
    \label{fig:enter-label}
\end{figure}
\begin{figure*}[htbp]
    \centering
    \includegraphics[width=\linewidth]{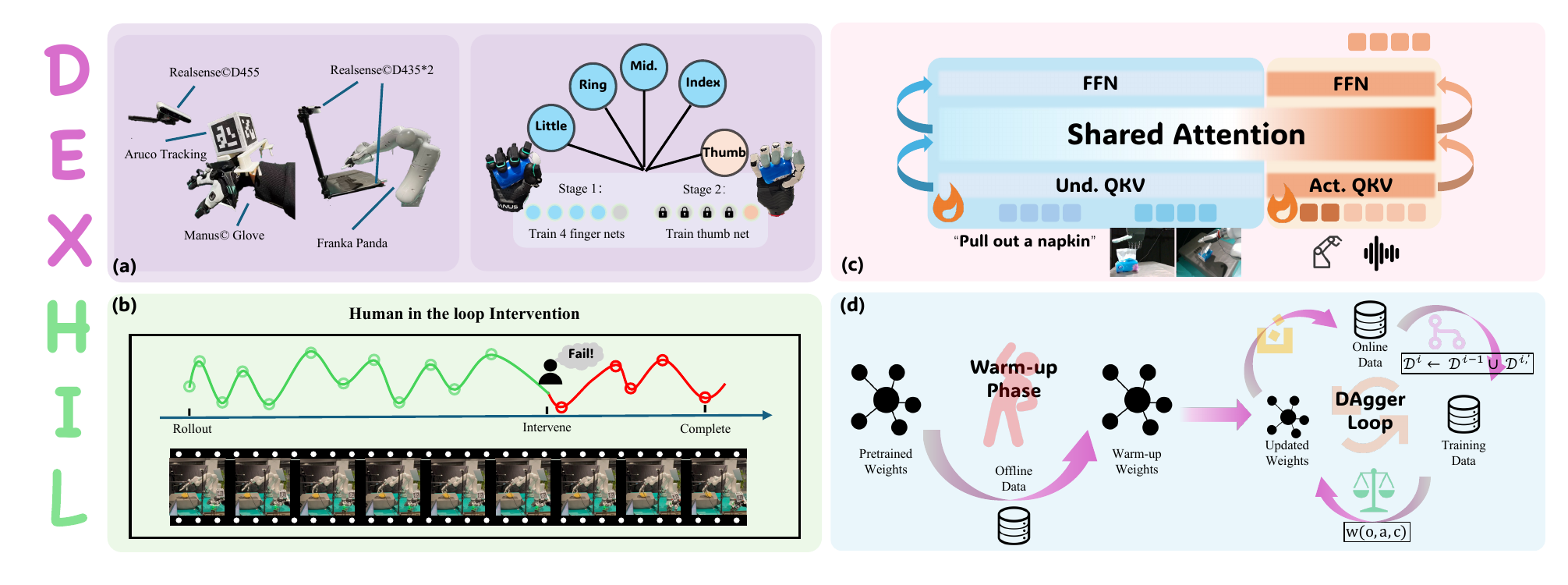}
    \caption{\textbf{The DexHiL Framework.} Below we will introduce our overall framework of DexHiL from data acquisition system, human-in-the-loop intervention paradigm, dexterous manipulation VLA model structure to offline-to-online training process. (\textcolor{violet}{a}) We propose an arm-hand data collection system both supporting teleoperation offline data collection and online human-in-the-loop policy intervention data collection. We also propose a two-stage training method for precise hand joint retargeting. (\textcolor{green}{b}) We propose an asynchronous human-in-the-loop policy intervention mechanism for online data collection, here we examplify the ``Plush Toy Grasping" case. (\textcolor{pink}{c}) Our dexterous manipulation VLA policy follows Being-H0.5~\cite{luo2026being} structure, which utilizing MoT (Mixture of Transformer) to relate understanding model with action expert for multi-modal reasoning and action generation and inherit the open-source pretrained weights of it. (\textcolor{blue}{d}) We propose a two-phase training framework that ultilize both offline dataset and online dataset. In the first warm-up phase, we finetuned the pretrained weights into warm-up model. In the DAgger loop, we utilize the system above to acquire online dataset and use reweighting training to update the policy which will be applied in the next DAgger loop.} 
    \label{fig:framework}
\end{figure*}
Beyond hardware limitations, the algorithmic post-training process for dexterous VLAs faces three core systemic challenges:
(i) Convergence difficulties in high-dimensional action spaces: The expansive action manifold of dexterous hands, compounded by complex contact dynamics, makes stable policy convergence exceptionally difficult;
(ii) Sample efficiency bottlenecks: Offline datasets are frequently dominated by repetitive success data, forcing the model to expend gradients on consolidating learned behaviors rather than exploring the critical transitions necessary for complex tasks;
(iii) Covariate shift and error accumulation: During real-robot execution, open-loop VLA policies suffer from trajectory drift. Without an effective recovery mechanism, minor errors quickly push the system into Out-of-Distribution (OOD) states, leading to failure~\cite{ross2011reduction}.

These challenges collectively suggest that relying solely on conventional offline post-training is insufficient for robust dexterous manipulation. To bridge this gap, we propose DexHiL, an interactive online training framework that integrates dexterous robotic hands with Human-in-the-Loop (HiL) intervention~\cite{kelly2019hg,liu2025robot}. DexHiL leverages a human-robot collaboration mechanism to facilitate precise manipulation and rapid error recovery in complex, contact-rich tasks. By incorporating real-time expert interventions and an intervention-aware weighting mechanism, DexHiL significantly enhances sample efficiency and policy convergence, ensuring robust performance in high-DOF real-world applications.

Our experimental results demonstrate that for a VLA model pre-trained on large-scale human videos with a cold-start action head, jointly fine-tuning on multiple downstream tasks using the DexHiL framework yields significant performance gains. Through three iterative stages of online refinement, DexHiL achieves 20\% and 30\% improvements in success rates across two distinct tasks, respectively, compared to a baseline model trained offline with an equivalent volume of data. Ablation studies further validate the necessity of our core components, demonstrating that the intervention-aware weighting mechanism is the primary driver for overcoming sample efficiency bottlenecks in high-DOF manipulation.

Our contributions are summarized as follows:
\begin{itemize}
\item \textbf{Human-to-Robot Hand Movement Retargeting}: We introduce a novel learning-based approach for precise retargeting of human hand movements to dexterous robotic hands. This method effectively maps human hand gestures to the robotic manipulator with high fidelity, ensuring accurate control over complex dexterous hand movements. Unlike traditional optimization methods, our approach provides a more adaptive, real-time mapping, improving the performance of dexterous manipulation tasks.

\item \textbf{Integrated and HiL-Enabled Teleoperation System}: We propose a seamless human-in-the-loop Teleoperation System that addresses the challenge of intervention discontinuity in high-DOF dexterous hand manipulation. By enabling smooth real-time interventions during post-training, our framework ensures effective and high-quality error correction.

\item \textbf{Iterative Human-in-the-Loop Post-Training for VLA}: We propose DexHiL, a post-training pipeline for dexterous VLA models that introduces a novel intervention-aware data sampling strategy. This strategy prioritizes corrective segments from expert interventions, ensuring that the most valuable data for error recovery and task refinement is efficiently used. By dynamically re-weighting these corrective samples, DexHiL accelerates convergence, improves sample efficiency, and enhances model performance, particularly in high-DOF, contact-intensive tasks.

\end{itemize}

\section{Related work}

\subsection{Dexterous Manipulation Data Collection System}

Dexterous manipulation data collection systems typically employ two approaches. The first method directly collects human hand data and generates dexterous hand data in a simulation environment using optimization methods~\cite{pan2025spider} or reinforcement learning~\cite{li2025maniptrans,xu2025dexplore,mandi2025dexmachina}.
The other relies on teleoperation with retargeting algorithms, typically utilizing exoskeletons or master-slave arms~\cite{wang2024dexcap,ben2025homie}. However, both approaches face limitations when applied to Human-in-the-Loop systems. Exoskeletons or master-slave arms do not align precisely with the current dexterous hand's robotic arm, preventing the intervention process from being made in an incremental form. Additionally, teleoperation methods often rely on optimization~\cite{handa2020dexpilot,wen2025dexterous,du2025mile} or network fitting~\cite{yin2025geometric,chong2021learning,handa2020dexpilot}, with the former struggling to control finger movement precisely, especially in grasping tasks, and the latter typically focusing on the thumb, leading to poor correspondence with the remaining fingers. Our framework addresses these issues by providing more accurate data collection and dexterous manipulation solutions.

\subsection{Vision Language Action model for dexterous manipulation}

Recent advancements in dexterous manipulation, exemplified by various frameworks, highlight the transformative potential of VLA models~\cite{luo2026being, li2025scalable} and Vision-Language Grasp(VLG) models~\cite{zhong2025dexgraspvla, he2025dexvlg}. However, these approaches typically require the collection of offline real-world datasets, followed by fine-tuning to adapt to downstream robots and tasks. In the high-dimensional action spaces inherent to dexterous hands, such offline post-training strategies often encounter significant distribution shifts and suffer from low sample efficiency~\cite{chen2025conrft}. These challenges make achieving convergence, particularly for contact-intensive tasks, especially difficult. To overcome these limitations, we propose an online post-training framework that mitigates these issues.

\subsection{Human-in-the-Loop Corrections for Robot Learning}
Interactive correction learning aims to rectify robotic behaviors through real-time human intervention~\cite{ross2011reduction,chen2025conrft,zhang2024diffusion}. 
Interactive correction learning refines robot behaviors through real-time human intervention~\cite{ross2011reduction,kelly2019hg,sun2023mega,liu2025robot}. HG-DAgger \cite{kelly2019hg} and its subsequent work, Sirius \cite{liu2025robot}, incorporate human corrections during critical moments to enhance policy efficiency~\cite{menda2019ensembledagger,hoque2021thriftydagger}. While these human-in-the-loop data aggregation paradigms have proven effective, they are predominantly confined to parallel gripper-arm setups and have not been successfully extended to dexterous manipulation~\cite{hoque2021thriftydagger,menda2019ensembledagger,cui2025end}. 
A recent work closely related to ours is DexGrasp-VLA \cite{cui2025end}, which applies human-in-the-loop correction only to the robotic arm. However, it does not integrate HIi for the hand, leaving hand movements controlled by a separate grasping network. This lack of coordination between the arm and hand limits the effectiveness of the data, especially for contact-rich tasks.

\section{Method}

We present DexHiL, a framework designed to tackle the complexities of high-dimensional control and data inefficiency in dexterous VLA. The methodology comprises two synergistic components: (1) an Interactive Human-in-the-Loop Teleoperation System for Dexterous Manipulation, which provides a unified interface for seamless arm-hand coordination and real-time intervention; and (2) a Human-in-the-Loop Post-training Pipeline, which leverages an intervention-aware weighting mechanism to accelerate policy convergence and mitigate covariate shift during real-world execution. Together, these components enable the model to efficiently learn robust error-recovery behaviors in contact-rich scenarios.
\subsection{Interactive Human-in-the-Loop Teleoperation System for Dexterous Manipulation}

\label{sec:teleop}

To support high-fidelity offline data collection and intuitive human-in-the-loop operation, we implement a lightweight arm--hand teleoperation interface. A handheld ArUco marker cube is tracked by a monocular camera to recover its 6D pose in real time, providing a low-overhead and fast-to-deploy solution that remains robust across setups. To account for the kinematic mismatch between the human operator and the robot, we use a \textbf{dual-path mapping}: one path aligns end-effector pose and global arm motion, while the other retargets finger articulations to the dexterous hand, jointly maintaining trajectory consistency and gesture-level controllability.

\subsubsection{Hand Joint Retargeting} 

We propose a unified hand-joint retargeting pipeline for dexterous hands with coupled joints. Human keypoints $\mathbf{X}_{\mathrm{keypoints}}$ are captured by a motion-capture glove, and the retargeting network $f_{\theta}$ maps them to the robot’s actuated joint angles $\mathbf{X}_{\mathrm{act}}\in\mathbb{R}^{m}$. The full joint configuration $\mathbf{q}$ is obtained by applying the hand’s predefined coupling constraints to $\mathbf{X}_{\mathrm{act}}$, and the robot fingertip positions are computed by forward kinematics as $\mathbf{P}_{\mathrm{tip}}=\mathrm{FK}(\mathbf{q})$.

To avoid the degenerate behavior observed in single-network five-finger retargeting, where the learned grasps collapse to pinch-like postures, we adopt a two-stage scheme.Empirically, optimizing all five fingers in a single network degrades the solution space of the four non-thumb fingers: the learned behaviors tend to concentrate on fingertip-to-thumb contacts, yielding stable opposition while the resulting ``grasps'' become pinch-like and fail to preserve enveloping grasp capability. In the first stage, we optimize only the index, middle, ring, and little fingers to obtain a stable and complete four-finger motion manifold.Supervision is imposed through intra-finger geometric constraints, which capture both articulation direction and extension length without relying on global five-finger couplings. Specifically, for each $i\in\mathcal{I}$, let $\mathbf{r}_i^{H}, \mathbf{r}_i^{R}(\mathbf{q})\in\mathbb{R}^3$ denote the human/robot root-to-tip vectors, with $d_i=\|\mathbf{r}_i^{H}\|_2$ and $\hat{\mathbf{r}}_i^{H}=\mathbf{r}_i^{H}/(d_i+\epsilon)$. We train with
\begin{equation}
\mathcal{L}_{\mathrm{vec}}
=\frac{1}{2}\sum_{i\in\mathcal{I}} s(d_i)\,
\big\|\mathbf{r}^{R}_i(\mathbf{q}) - f(d_i)\,\hat{\mathbf{r}}^{H}_i\big\|_2^{2}
+\gamma\|\mathbf{q}\|_2^2 ,
\label{eq:kin_vec}
\end{equation}
where $s(\cdot)$ and $f(\cdot)$ are distance-dependent weight and scale functions. 

After training the non-thumb fingers, we freeze their retargeting parameters and optimize only a thumb residual mapping. This stage aims to preserve the mapping from the human fingertip C-space to the robot fingertip C-space. Following the objective design of GeoRT~\cite{yin2025geometric}, we optimize the thumb residual using a concise set of geometric regularizers, including the motion-preservation loss $\mathcal{L}_{\mathrm{dir}}$, workspace-coverage loss $\mathcal{L}_{\mathrm{cover}}$, flatness loss $\mathcal{L}_{\mathrm{flat}}$, and pinch-preservation loss $\mathcal{L}_{\mathrm{pinch}}$.We further introduce an inter-fingertip kinematic term $\mathcal{L}_{\mathrm{kin}}$, which applies Eq.~\eqref{eq:kin_vec} to a predefined set of five-finger fingertip-pair vectors (thumb--finger and finger--finger) to match human and robot inter-finger geometry.We train the model with the following combined objective:
\begin{equation}
\begin{aligned}
\mathcal{L}
=&~\lambda_{\mathrm{dir}}\mathcal{L}_{\mathrm{dir}}
+\lambda_{\mathrm{cover}}\mathcal{L}_{\mathrm{cover}}
+\lambda_{\mathrm{flat}}\mathcal{L}_{\mathrm{flat}} \\
&+\lambda_{\mathrm{pinch}}\mathcal{L}_{\mathrm{pinch}}
+\lambda_{\mathrm{kin}}\mathcal{L}_{\mathrm{kin}} .
\end{aligned}
\label{eq:stage2_loss}
\end{equation}

\subsubsection{Arm Pose Mapping} 
Let $\mathbf{T}_{M} \in SE(3)$ denote the 6-DoF pose of the ArUco cube expressed in the camera frame. Assuming the constant transformation matrix from the cube frame to the robot's end effector (EE) frame is known and indicated as $\mathbf{T}_{\mathrm{robot}}^{\mathrm{cube}}$. During policy rollout, the human operator requires a key press on the keyboard to trigger the mechanism of \textbf{human intervention}. Upon activation, we will record the current robot EE pose $\mathbf{T}_{EE_0}$ and ArUco cube pose $\mathbf{T}_{M_0}$ as \textbf{anchor poses}. To enable a smooth and intuitive takeover, we map the cube’s subsequent motion to the robot by applying the cube’s relative pose (w.r.t. the anchor) to the EE anchor pose. Hence, the target pose of the robot's EE in the base frame, $\mathbf{T}_{EE}$, is computed as:
\begin{equation}
    \mathbf{T}_{EE} = \mathbf{T}_{EE_0}(\mathbf{T}_{\mathrm{robot}}^{\mathrm{cube}})^{-1}\mathbf{T}^{-1}_{M_0}\mathbf{T}_{M}\mathbf{T}_{\mathrm{robot}}^{\mathrm{cube}}
\end{equation}

The corresponding arm joint angles $\mathbf{q}_{arm} \in \mathbb{R}^n$ are then derived via inverse kinematics (IK):
\begin{equation}
    \mathbf{q}_{arm} = \mathcal{K}^{-1}(\mathbf{T}_{EE})
\end{equation}

This marker-based technique provides high robustness to ambient lighting and efficient detection even in cluttered environments

\subsubsection{Asynchronous Multi-threaded Control and Intervention}

We also designed a multi-threaded architecture to handle autonomous execution and human intervention concurrently. The autonomous policy $\pi$ performs inference at 20Hz, while the human-guided arm and hand teleoperation operate at 30Hz and 90Hz, respectively. The human operator monitors the robot's performance and takes over the execution process of the entire robot system upon detecting imminent task failure.

The resulting dataset is formally represented as:
\begin{equation}
    \mathcal{D} = \{ (\mathbf{o}_t, \mathbf{q}_{arm,t}, \mathbf{q}_{hand,t}, I_t) \}_{t=1}^T\label{eq:dataset}
\end{equation}
where $I_t \in \{0, 1\}$ is a binary indicator. The control law $u_t$ at time $t$ can be described as:
\begin{equation}
    u_t = 
    \begin{cases} 
    \pi(\mathbf{o}_t), & \text{if } I_t = 0 \text{ (Autonomous)} \\
    u_{human}, & \text{if } I_t = 1 \text{ (Intervention)}
    \end{cases}
\end{equation}
where $u_{human}$ is the command derived from the mapping described in Eqs. (1)-(3).
\begin{figure*}[t]
    \centering
    \includegraphics[height=8cm,width=\linewidth]{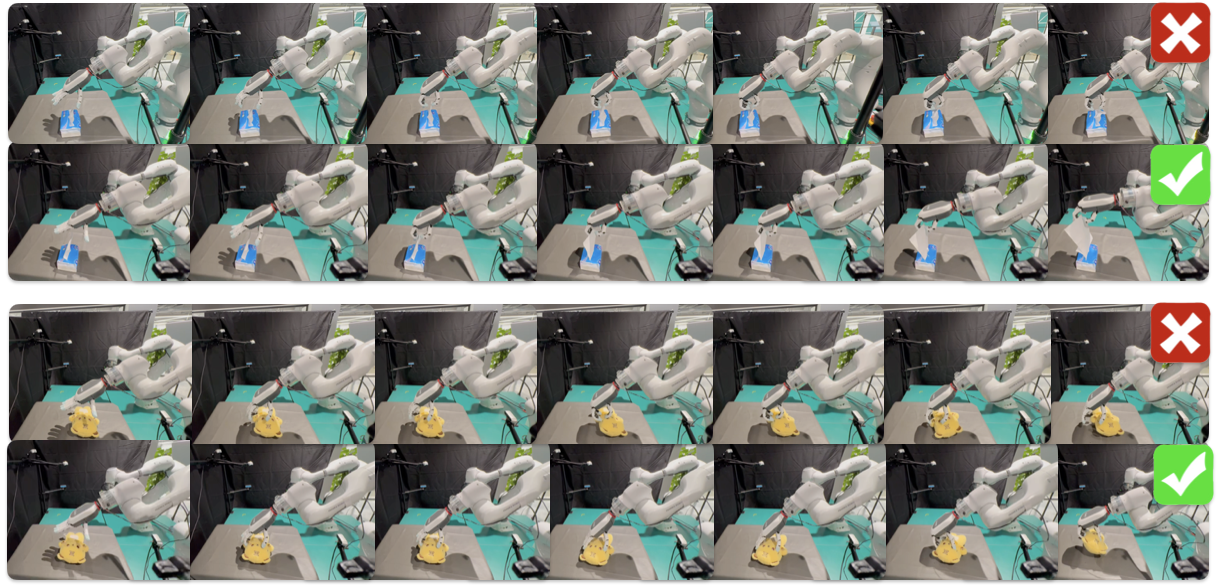}
    \caption{\textbf{Real-world rollouts of dexterous manipulation tasks.} (Up) \textbf{Tissue Extraction:} The system achieves precise fingertip alignment and vertical retraction to extract the tissue. (Down) \textbf{Plush Toy Grasping:} The controller executes a synchronized multi-joint flexion to securely envelop and lift the deformable object.}
    \label{fig:rollout}
\end{figure*}
\subsection{Human-in-the-Loop Post-training pipeline}

\label{sec:hil}

We propose a Human-in-the-Loop interactive post-training pipeline for VLA models, built upon the teleoperation algorithm and hardware system described in Sec.~\ref{sec:teleop} (Fig.~\ref{fig:framework}). The framework incorporates a policy refinement mechanism and a intervention-aware weighting training strategy that combines both online human-in-the-loop data and offline datasets, enhancing the ability of Human-in-the-Loop to improve VLA models for dexterous manipulation tasks.

\subsubsection{Intervention-aware Weighting Mechanism}
Prior work has demonstrated that human intervention data is crucial for robot learning \cite{liu2025robot}, enabling models to learn error avoidance and recovery from sub-optimal actions. However, such data is often sparse compared to offline datasets. To improve the utilization of these high-value samples, we use a intervention-aware weighting mechanism, which incorporates the weighting term $w(o,a,c)$ into our policy update framework.

Let the aggregated dataset $\mathcal{D}$ contain $N$ samples, where $n_c$ is the number of samples in category $c$. The original empirical distribution is defined as $P(c) = n_c / N$. The intervention-aware weighting mechanism adjusts this to a target distribution $P^*(c)$ with an increased proportion of interventions. Based on the principle of importance sampling, the sample weight $w(o,a,c)$ for category $c$ is:
\begin{equation}
    w(o,a,c) = \frac{P^*(c)}{P(c)}
\end{equation}
To enable the model to more effectively master contact-rich maneuvers that are otherwise prone to failure, we specifically re-weight the sparse intervention trajectories. In our implementation, we specify $P^*(\text{intervention}) = 0.5$, an empirical equilibrium that further functions as a regularization mechanism, bridging the distribution shift without compromising foundational capabilities. Ultimately, this approach leads to enhanced deployment robustness and superior sample efficiency.

\subsubsection{Policy Update Mechanism}
The policy refinement process is structured into three integrated stages: warm-up initialization, online interactive learning, and strategy-driven data filtering.

\paragraph{Warm-up Phase}
We begin with a warm-up phase, where we collect an initial offline dataset $\mathcal{D}^{0}$ via our teleoperation system. The VLA model undergoes full fine-tuning on $\mathcal{D}^{0}$ to derive the initial robotic policy $\pi_{0}$:

\begin{equation}
\theta^{(0)}=\arg\min_{\theta}\ \mathcal{L}(\theta;\mathcal{D}^{0}),\qquad
\pi_{0}=\pi_{\theta^{(0)}}.
\end{equation}

where $\theta^{(0)}=(\theta_{\mathrm{VLM}},\theta_{\mathrm{A}})$ denotes the whole parameters comprised with VLM and action head parameters, warmed-up with $\mathcal{D}^{0}$. 

\paragraph{Online Training Phase}
Upon obtaining $\pi_{0}$, the system enters an online learning loop for $n$ iterations. During each iteration $i \in \{1,2,\dots,n\}$, the current policy is deployed on the physical robot under human supervision. When an imminent failure is detected, the human intervenes to provide corrective demonstrations, yielding an online dataset $\mathcal{D}^{i,\prime}$. Following dataset aggregation, we form
\begin{equation}
\mathcal{D}^{i} \leftarrow \mathcal{D}^{i-1} \cup \mathcal{D}^{i,\prime}, \quad i=1, 2, \dots, n
\end{equation}

We then update the policy by warm-starting from the previous parameters $\theta^{(i-1)}$ and performing supervised post-training on $\mathcal{D}^{i}$. Concretely, we optimize a weighted imitation objective:

\begin{equation}
\mathcal{L}^{(i)}(\theta;\mathcal{D}^{i})
= \mathbb{E}_{(o,a,c)\sim \mathcal{D}^{i}}
\Big[w(o,a,c)\cdot \ell_{\mathrm{IL}}(\theta; o,a)\Big],
\label{eq:online_obj}
\end{equation}

where $w(o,a,c)$ is instantiated by our intervention-aware weighting mechanism to emphasize intervention samples, and $\ell_{\mathrm{IL}}$ denotes the per-sample imitation loss.

In this paper, we instantiate DexHiL with Being-H0.5~\cite{luo2026being} as a representative VLA model; since its action head follows a Flow Matching (FM) formulation, we use the FM objective for $\ell_{\mathrm{IL}}$:
\begin{equation}
\ell_{\mathrm{IL}}(\theta; o,a)
=
\mathbb{E}_{t, x_t}\left[
\left\| v_\theta(x_t,t,o) - u_t(a \mid x_0) \right\|_2^2
\right],
\label{eq:fm_loss_online}
\end{equation}
where $o$ represents the multimodal observation, $a$ is the expert action, and $x_t = (1-t)x_0 + ta$ is the probability path interpolated between Gaussian noise $x_0$ and the target action $a$. The model $v_\theta$ learns to predict the ideal velocity field $u_t(a|x_0) = a - x_0$. The integration of sample weights $w(o,a,c)$ ensures that the VLA model prioritizes gradients derived from high-value intervention data.

The parameters are updated via stochastic gradient descent:
\begin{equation}
\theta^{(i)} \leftarrow \theta^{(i-1)} - \eta \nabla_\theta \mathcal{L}^{(i)}(\theta;\mathcal{D}^{i}),
\qquad
\pi_i = \pi_{\theta^{(i)}},
\label{eq:sgd_update}
\end{equation}
with learning rate $\eta$. This iterative human-in-the-loop refinement progressively improves the policy’s robustness, especially in failure-prone contact-rich manipulation scenarios.

\paragraph{Data Filtering Strategy}

\label{sec:filter}

To enhance online optimization, we implement a targeted data filtering scheme. For each trial involving human interventions, we record data according to \eqref{eq:dataset}, but preserve only the segments from the final intervention to task completion, while discarding all trajectory segments prior to the last takeover. 

The rationale behind this filtering is that trajectories with multiple interventions often exhibit significant action incoherence. Furthermore, the model actions preceding the intervention are inherently sub-optimal or erroneous \cite{liu2025robot}. Training on such inconsistent data can lead to Policy Oscillation or Multimodal Distribution Conflict. By focusing on the terminal recovery segments, the model adopts a Progressive Error Correction paradigm, which effectively facilitates the learning of robust manipulation skills within high-dimensional action spaces.

\section{EXPERIMENTS}

In this section, we conduct comprehensive experiments to validate our method from three complementary perspectives:
\textbf{RQ1:} How does DexHiL efficiently improve the performance of the dexterous hand VLA policy? Can the system improve the model's performance over time?\\
\textbf{RQ2:} To what extent do our hardware-specific algorithmic designs contribute to the overall performance in complex manipulation tasks?
\subsection{Experimental Setup}
\subsubsection{Tasks}
To demonstrate the capabilities of our method, we design tasks that emphasize two complementary aspects. First, we evaluate the overall effectiveness of the full system, providing an intuitive validation of the proposed HiL framework. Second, we assess the advantages of our modular design by testing whether it can accomplish fine-grained, dexterous-hand-specific manipulation tasks. We consider:

\begin{itemize}
  \item \textbf{Plush toy grasping.} The dexterous hand grasps and lifts a plush toy from a tabletop. This task primarily reflects the end-to-end policy performance and the hand’s grasping capability.
  \item \textbf{Tissue extraction.} The dexterous hand pulls a single tissue from a tissue pack. This is a more challenging task that requires sufficiently precise dexterous retargeting to reliably extract a thin, deformable tissue.
\end{itemize}
The execution sequences for both tasks are depicted as sequential real-world rollouts in Fig.~\ref{fig:rollout}.

\subsubsection{Implementation Details} 
Our robotic platform comprises a Franka Research 3 arm equipped with a DexHand021 dexterous hand. Initially, we use 60 offline trajectories for model initialization.In subsequent rounds, we expand the training set by adding 10 additional trajectories per task per round. We consider two conditions: (i) trajectories newly collected in-the-loop via our Dex-HiL procedure, and (ii) a non-HiL baseline where we instead add 10 trajectories per round from the remaining offline pool, matching the data volume while removing the benefit of interactive collection. This protocol isolates the effect of HiL-based post-training under the same data budget.

For training, we perform full training of \textbf{Being-H0.5} on 8 NVIDIA H100 GPUs for 60k training iterations, and then fine-tune on the human-interaction data using a single H100 GPU.

\begin{figure}[htbp]
    \centering
    \includegraphics[width=0.5\textwidth]{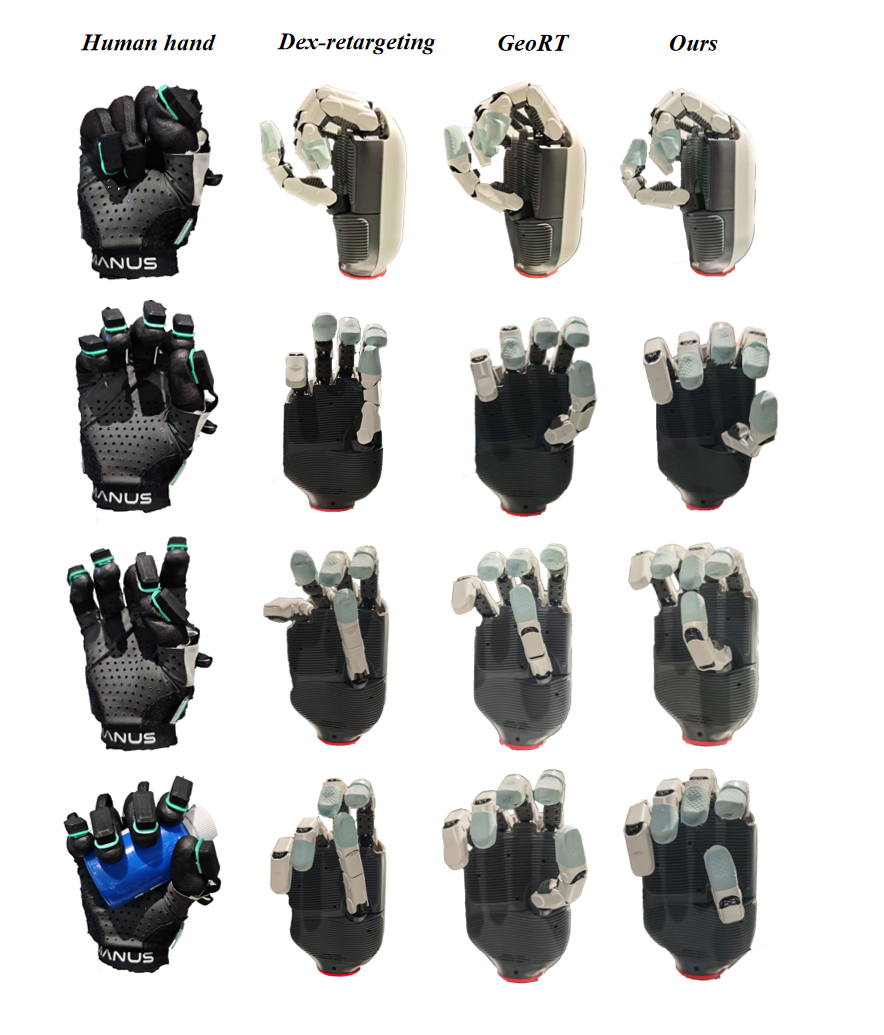} 
\caption{\textbf{Visualization of retargeting results for four representative gestures.} We show the input \textbf{human hand} poses and the corresponding configurations generated by \textbf{Dex-retargeting~\cite{qin2023anyteleop}}, \textbf{GeoRT~\cite{yin2025geometric}}, and \textbf{our method}.Compared to other methods, our retargeting algorithm generates more accurate, smooth, and coordinated hand poses.}
\label{fig:hand_results}
\end{figure}
\subsection{Baselines and Evaluation Protocol}

To evaluate the efficacy of our framework, we design a \textit{data-budget-matched} protocol for baseline comparison. During the fine-tuning of \textbf{Being-H0.5}, specific language instructions are utilized: ``\textit{Grab that yellow plushie}'' for grasping and ``\textit{Pull out a napkin}'' for extraction. All baselines are trained via full-parameter fine-tuning to ensure consistency.

\textbf{Offline-40 (Data-matched R1):} Employs 40 offline trajectories per task (80 total), matching the data budget of our HiL method at the first iteration round.

\textbf{Offline-50 (Data-matched R2):} Employs 50 offline trajectories per task (100 total), matching the data budget of our HiL method at the second iteration round.

\textbf{Offline-60 (Data-matched R3):} Employs 60 offline trajectories per task (120 total), matching the data budget of our HiL method at the third iteration round.

\textbf{Evaluation Protocol:} We evaluate the policies on two challenging dexterous tasks with rigorous success criteria. For \textit{Tissue extraction}, a trial is successful if the extracted length exceeds half of the napkin's length. For \textit{Plush toy grasping}, success is defined as the object being lifted entirely off the tabletop. We conduct 20 independent trials for each task on the physical robot and report the \textbf{Success Rate} as the primary metric. All experiments are initialized with the same pre-trained weights to ensure a fair comparison.

\subsection{Experiment Results}
\begin{figure*}[htbp]
    \centering
    \includegraphics[height=6cm,width=\linewidth]{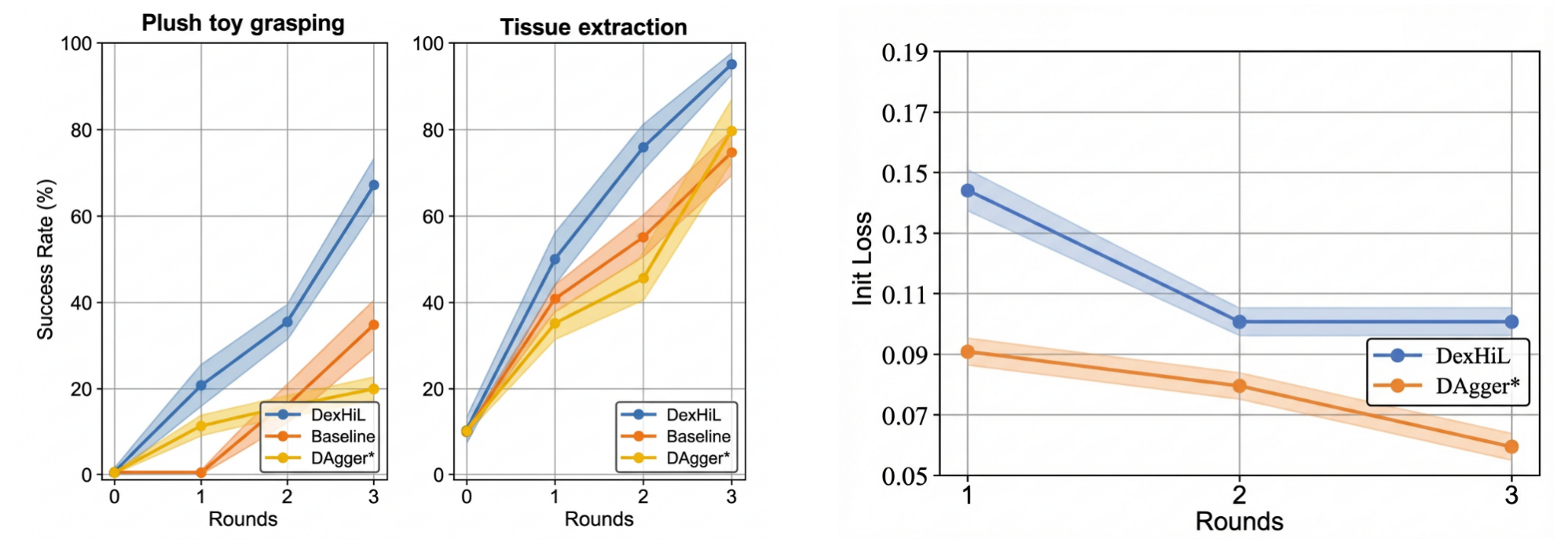}
    \caption{\textbf{Quantitative performance and training loss analysis.} (Left) \textbf{Success rates} across three consecutive training rounds for both Tissue Extraction and Plush Toy Grasping tasks. (Right) \textbf{Initial training loss} at step 10 for both DexHiL and DAgger*. While the loss previously plateaued between 0.002 and 0.008, it significantly increases after incorporating human corrective trajectories.}
    \label{fig:result_line}
\end{figure*}

Across both tasks, DexHiL consistently outperforms the baselines as training rounds progress (Tables~\ref{tab:combined_results}). In \textit{Tissue Extraction}, DexHiL achieves a 95\% success rate by Round 3, surpassing DAgger* (80\%) and the offline Baseline (75\%). Similarly, in \textit{Plush Toy Grasping}, DexHiL reaches 65\% success, whereas DAgger* and the Baseline struggle to scale, concluding at only 20\% and 35\%, respectively. These results demonstrate the efficacy of our intervention-aware mechanism in facilitating policy refinement through human corrections.
Beyond performance, DexHiL exhibits higher efficiency: each intervention segment takes only ~3s (vs. 10s for offline), yielding a 35\% reduction in total human labor (13 min vs. 20 min) by Round 3. This proves that our intervention-aware mechanism effectively prioritizes high-value corrective signals over redundant data.

\begin{table}[htbp]
    \centering
    \begin{threeparttable}
        \caption{\textbf{Success Rates ($n/20$)}}
        \label{tab:combined_results}
        
        \normalsize 
        \setlength{\tabcolsep}{2pt}
        \renewcommand{\arraystretch}{1.3} 
        
        \begin{tabular}{lcccc}
            \toprule
            \textbf{Method} & \textbf{Warm-up} & \textbf{Round 1} & \textbf{Round 2} & \textbf{Round 3} \\
            \midrule
            \multicolumn{5}{l}{\textit{Task 1: Tissue Extraction}} \\ 
            \midrule 
            DexHiL (Ours)    & 2/20 & \textbf{10/20} & \textbf{15/20} & \textbf{19/20} \\
            DAgger*\tnote{1} & 2/20 & 7/20 & 9/20 & 16/20 \\
            Baseline\tnote{2} & 2/20 & 8/20 & 11/20 & 15/20 \\
            
            \addlinespace[1.5ex] 
            
            \multicolumn{5}{l}{\textit{Task 2: Plush Toy Grasping}} \\
            \midrule
            DexHiL (Ours)    & 0/20 & \textbf{4/20} & \textbf{6/20} & \textbf{13/20} \\
            DAgger* & 0/20 & 2/20 & 3/20 & 4/20 \\
            Baseline         & 0/20 & 0/20 & 3/20 & 7/20 \\
            \bottomrule
        \end{tabular}
        
        \begin{tablenotes}
            \footnotesize
            \item[1] \textbf{DAgger*}: Online training without intervention-aware mechanism.
            \item[2] \textbf{Baseline}: Control group with same number of trajectories.
        \end{tablenotes}
    \end{threeparttable}
\end{table}
In embodied dexterous control, intervention timing is pivotal. Our experiments identify two dominant failure modes that bottleneck performance: (1) \textbf{suboptimal contact-rich maneuvers}, such as failing to pinch a tissue edge, and (2) \textbf{arm-hand discoordination}, characterized by premature grasping before the wrist reaches the target. DexHiL addresses these by triggering human corrections at \textit{impending failures} and preserving the subsequent recovery trajectories. This strategy captures essential state transitions required for error recovery, leading to significant performance gains. For instance, in \textit{Tissue Extraction}, mastering the precise pinching maneuver enables the success rate to scale rapidly, nearly reaching a perfect ceiling by Round 3. Similarly, in the coordination-intensive \textit{Plush Toy Grasping} task, DexHiL maintains a strong upward trajectory, substantially outperforming DAgger* and the unweighted baseline, which both struggle to overcome the complexity of the coordination bottleneck.

Beyond performance ceilings, DexHiL significantly improves sample efficiency. As illustrated in the training dynamics (Fig.~\ref{fig:result_line}, right), the updated policy $\pi_k$ exhibits distinct loss spikes at the onset of each iteration ($k=1,2,3$). These spikes signify a substantial \textit{distribution shift}, where human corrections introduce critical out-of-distribution states and compounding errors that the nominal policy initially fails to handle. While standard methods like DAgger* dilute learning capacity across the entire dataset, our intervention-aware weighting mechanism prioritizes optimization on these high-value recovery states. By extracting more informative gradients per demonstration, DexHiL enables the policy to converge to expert-level success rates in fewer interaction rounds, demonstrating superior convergence speed compared to standard intervention paradigms.

To answer RQ2, we compare our method with two representative dexterous-hand teleoperation approaches discussed in the paper. As shown in Fig.~\ref{fig:hand_results}, the optimization-based Dex-Retargeting~\cite{qin2023anyteleop} is constrained by hard, threshold-like control behavior: it tends to initiate finger opposition before the fingertips have established contact, leading to discontinuous adjustments around contact and making it difficult to smoothly regulate the hand configuration during grasp formation. In contrast, the learning-based GeoRT~\cite{yin2025geometric} struggles to form stable grasp postures: except for the thumb, the other four fingers often fail to fully flex to the intended closure, which makes palmar grasps that rely on coordinated contact between the palm and the four fingers difficult to execute reliably. By overcoming these limitations in control continuity and grasp stability, it is precisely our algorithmic design that enables the high-precision finger control required for tasks like Tissue Extraction, where DexHiL achieves a 95\% success rate.

\section{CONCLUSION AND FUTURE WORK}

We introduced DexHiL, an arm–hand integrated human-in-the-loop post-training framework for dexterous VLA. By combining a lightweight teleoperation interface with intervention-aware sampling that emphasizes corrective segments, DexHiL turns online interventions into high-value post-training data for contact-rich, high-DOF manipulation. Real-robot experiments demonstrate consistent gains in task success, establishing DexHiL as an effective and practical post-training recipe for improving dexterous VLA policies. After completing ongoing hardware and teleoperation optimizations, we will study dexterous-hand representation in VLA—particularly hand tokenizers and integrate them with our post-training pipeline to further improve performance and generalization.

\bibliographystyle{IEEEtran}
\bibliography{IEEEtranBST/IEEEexample}

\end{document}